# Cascaded Structure Tensor Framework for Robust Identification of Heavily Occluded Baggage Items from Multi-Vendor X-ray Scans

Taimur Hassan*, Samet Akçay, Mohammed Bennamoun, , Salman Khan, Naoufel Werghi,

*Abstract*—In the last two decades, luggage scanning has globally become one of the prime aviation security concerns. Manual screening of the baggage items is a cumbersome, subjective and inefficient process. Hence, many researchers have developed X-ray imagery-based autonomous systems to address these shortcomings. However, to the best of our knowledge, there is no framework, up to now, that can recognize heavily occluded and cluttered baggage items from multi-vendor X-ray scans. This paper presents a cascaded structure tensor framework which can automatically extract and recognize suspicious items irrespective of their position and orientation in the multi-vendor X-ray scans. The proposed framework is unique, as it intelligently extracts each object by iteratively picking contour based transitional information from different orientations and uses only a single feedforward convolutional neural network for the recognition. The proposed framework has been rigorously tested on publicly available GDXray and SIXray datasets containing a total of 1,067,381 X-ray scans where it significantly outperformed the state-of-the-art solutions by achieving the mean average precision score of 0.9343 and 0.9595 for extracting and recognizing suspicious items from GDXray and SIXray scans, respectively. Furthermore, the proposed framework has achieved 15.78% better time performance as compared to the most popular object detectors.

*Index Terms*—Aviation Security, Baggage Screening, Convolutional Neural Networks, Image Analysis, Structure Tensor, X-ray Radiographs

## I. INTRODUCTION

X-ray imaging is a widely adapted tool in healthcare for the diagnosis and visualization of different medical conditions [1-2]. Apart from this, it is also utilized by various manufacturing industries for non-destructive testing (NDT), in particular for the baggage inspection at airports, malls and cargo transmission trucks [3]. Luggage threats have become the prime concern all over the world. According to a recent report, approximately 1.5 million passengers are searched every day in the United States against weapons and other dangerous items [4]. The manual detection of such items in each baggage is a cumbersome and time-consuming process. Therefore, aviation authorities, all over the world, are actively looking for automated and reliable baggage screening systems. Object detection and recognition within the computer vision community is a well-known problem [5-7]. Furthermore, a number of large-scale natural image datasets for object detection are publicly available (free of charge), enabling the development of popular object detectors like R-CNN [8], SPP-Net [9], YOLO [10] and RetinaNet [11] etc. In contrast, only few datasets for X-ray images are currently available for researchers to develop robust computer aided screening systems. Apart from this, the nature of radiographs is quite different than natural photographs. Although they can reveal the information invisible in the normal photographs (due to radiations), but they lack texture (especially the grayscale scans), due to which conventional detection methods do not perform well on them [12]. In general, screening objects and anomalies from baggage X-ray (grayscale or colored) scans is a challenging task especially when the objects are closely packed to each other leading to heavy occlusions. In addition to this, luggage screening systems faces severe class imbalance problem due to the low suspicious to normal items ratio. Therefore, it is highly challenging to develop an unbiased decision support system that can effectively screen baggage items despite the high contribution of the normal items within the training images. Fig. 1 shows some of the X-ray baggage scans where the suspicious items such as *guns* and *knives* are highlighted in a heavily occluded and cluttered environment. Several methods for classifying objects in X-ray imagery have been proposed. Most of these methods are based on key-point descriptors [12], bag of visual words [13], and handcrafted features [14]. The general applicability of these methods is limited, as they are tested and validated in a controlled environment on limited datasets.

This work is supported by Center for Cyber-Physical Systems (C2PS).
T. Hassan and N. Werghi are with the Center for Cyber-Physical Systems (C2PS), Department of Electrical Engineering and Computer Science, Khalifa University, Abu Dhabi UAE.
S. Akçay is with the Department of Computer Science, Durham University, Durham UK.

M. Bennamoun is with the Department of Computer Science and Software Engineering, The University of Western Australia, Perth Australia.
S. H. Khan is with the Inception Institute of Artificial Intelligence, Abu Dhabi UAE.
*Corresponding Author (e-mail: engr.taimoorhassan@gmail.com)

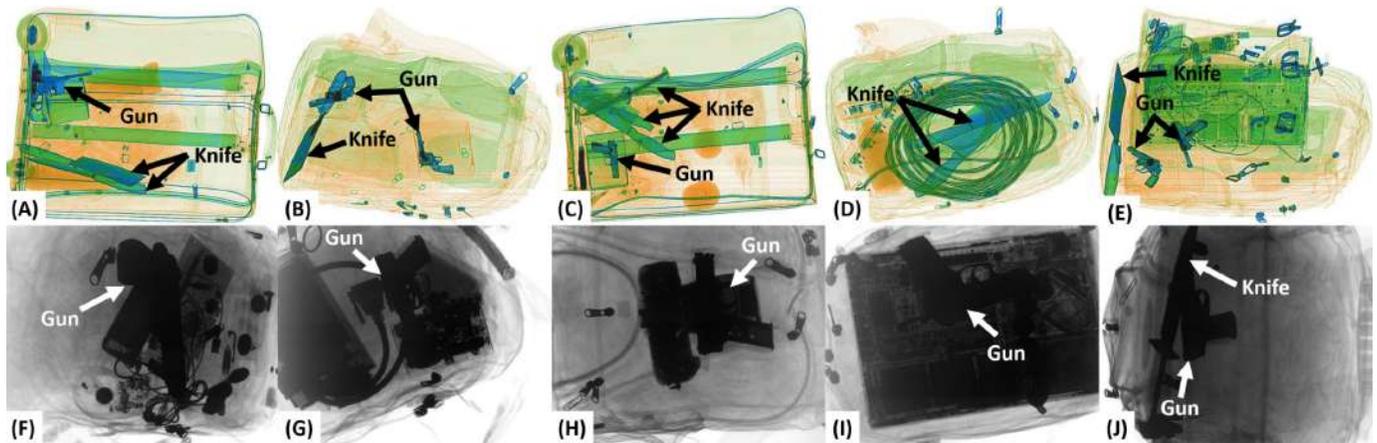

Fig. 1. Exemplar X-ray images showing heavily occluded and cluttered items. Top row shows scans from SIXray dataset [46] while bottom row shows scans from GDXray dataset [49].

## II. RELATED WORK

Since the inception of deep learning, computer vision field has progressed tremendously. This progress manifested in the employment of convolutional neural networks (CNNs) to detect objects from images and videos [15-21]. These methods have outperformed traditional handcrafted features-based approaches in terms of robustness and accuracy. Apart from this, we have also developed several CNN based methods for image classification [22], object detection [23], recognition [24] and proposed many loss function plugins for the deep networks based on Gaussian affinity measure [25] and cost-sensitive learning [26] to boost their performance on the highly imbalanced data. More recently, we have developed many supervised and unsupervised methods for real-time threat detection based on security X-ray imagery [27-32]. However, the robust identification of cluttered and heavily occluded baggage items from the complex multi-vendor X-ray scans is still an unsolved problem. The literature related to the suspicious items detection from X-ray images has been broadly categorized into traditional and deep learning approaches.

### A. Traditional Approaches

The development of automated tools to screen baggage items is not new and many researchers have used traditional machine learning (ML) approaches to recognize baggage items from the X-ray scans. Bastan et al. [13] proposed a structured learning framework that encompasses dual-energy levels for the computation of low textured key-points to detect *laptops*, *handguns* and *glass bottles* from multi-view X-ray imaging. They also presented a framework that utilizes Harris, SIRF, SURF and FAST descriptors to extract key-points which are then passed to a bag of words (BoW) model for the classification and retrieval of X-rays images [14]. They concluded that although BoW produces promising results on regular images, it does not perform well on low textured X-ray images [14]. Chen et al. [33] proposed a scheme that employs wavelet transform on dual-energy X-ray images for threat detection. Hanif et al. [34] proposed a framework that uses Wi-Fi signals for the non-obtrusive detection of metallic and non-metallic concealed items. Jaccard et al. [35] proposed an automated method for detecting cars from the X-ray cargo transmission images based upon their intensity, structural and symmetrical properties using a random forest classifier. Al-Zubi et al. [36] used SIFT descriptors to automatically detect concealed weapons from X-ray images. They prepared their own local dataset and applied their proposed method on it. Rogers et al. [37] presented a framework that adds threat image projections (TIP) into the X-ray images for training ML algorithms as well as humans for real threat detection. TIP also encompasses translation, magnification, rotation, noise and illumination effects. Turcsany et al. [38] proposed a framework that uses SURF descriptor to extract distinct features and passes them to a BOW for the object recognition in baggage X-ray images. Riffo et al. [39] proposed adapted implicit shape model (AISM) for the autonomous detection of threatening items in baggage X-ray imagery. Pourghassem et al. [40] used connected component analysis and extracted shape features which are passed to a probabilistic neural network classifier for weapon detection through dual-energy X-ray images.

### B. Deep Learning Based Suspicious Items Detection

Many researchers have recently presented studies in which deep learning architectures are employed for the detection and classification of suspicious baggage items from X-ray imagery. These studies are either focused on the usage of supervised classification models or an unsupervised adversarial learning:

#### 1) Unsupervised Anomaly Detection

Akcay et al. proposed GANomaly [27] and Skip-GANomaly [28] architectures to detect different anomalies from X-ray scans. These approaches employ an encoder-decoder for deriving a latent space representation used by discriminator network to classify anomalies. Both architectures are trained on normal distributions while they are tested on normal and abnormal distributions from CIFAR-10, Full Firearm vs Operational Benign (FFOB) and the local in-house datasets (GANomaly is also verified on MNIST dataset).

#### 2) Supervised Approaches

Gaus et al. [29] proposed a dual CNN based framework in which the first CNN model detects the object of interest and then the second CNN model classifies it as benign or malignant. For object detection, the authors performed an evaluation of Faster R-CNN, Masked R-CNN and RetinaNet for classification using VGG-16. Zou et al. [41] used a pre-trained YOLOv2 model to automatically identify suspicious items from X-rays images. They generated their own dataset containing 1,104 synthetic X-rays images of suspicious items including *scissors*, *knives* and *bottles*. Akcay et al. [30] used a pre-trained GoogleNet model for object classification from X-ray baggage scans. They prepared their in-house dataset and tested their proposed framework to detect *cameras*, *laptops*, *guns*, *gun components* and *knives* (particularly *ceramic knives*). Dhiraj et al. [42] used YOLOv2, Tiny YOLO and Faster R-CNN models to extract *guns*, *shuriken*, *razor-blades* and *knives* from baggage X-ray scans of GRIMA X-ray database (GDXray) where they achieved an accuracy of up to 98.4% and their proposed framework takes 0.16 seconds to process a single image. Liu et al. [43] used YOLO9000 for the autonomous detection of *scissors* and *aerosols* from X-ray images. Morris et al. [44] evaluated different CNN models to automatically identify threatening items via X-ray imagery. Their evaluations are conducted with a non-publicly available passenger baggage object database (PBOD). Xiao et al. [45] proposed an optimized form of Faster R-CNN (called R-PCNN) for terahertz (THz) imagery. R-PCNN reduced the training time of Faster R-CNN from 374 minutes to 150 minutes and the detection time from 37 milliseconds to 16 milliseconds. Akcay et al. [31] compared different frameworks for the object classification from X-ray imagery. They concluded that AlexNet as feature extractor with support vector machines (SVM) perform better than other ML methods. For occluded data, they compared the performance of sliding window-based CNN (SW-CNN), Faster R-CNN, region based fully convolutional networks (R-FCN) and YOLOv2 for object recognition. They used their local datasets as well as non-publicly available FFOB and Full Parts Operation Benign (FPOB) datasets in their experimentations. Miao et al. [46] provided one of the largest and challenging X-ray imagery datasets (named as SIXray) for detecting suspicious items. This dataset contains 1,059,321 scans with heavily occluded and cluttered objects in which 8,929 suspicious items are manually marked into six classes. Furthermore, the dataset handles the class imbalance problem of real-world scenarios by providing different subsets in which positive and negative samples ratios are varied [46]. The authors have also developed a deep class-balanced hierarchical refinement (CHR) framework that iteratively infers the image content through reverse connections and uses a custom class-balanced loss function to easily recognize the negative samples. The CHR framework was also validated on ILSVRC2012 large scale image classification dataset. After the release of SIXray dataset, Gaus et al. [32] evaluated Faster R-CNN, Mask R-CNN and RetinaNet on it as well as on other non-publicly available datasets.

To the best of our knowledge, all the methods which have been proposed in the past are either tested on single dataset or on the datasets containing similar type of X-ray imagery. Furthermore, there are limited frameworks which are applied on the complex radiographs for the detection of heavily occluded and cluttered baggage items. Many latest frameworks which can detect multiple objects and potential anomalies from the X-ray scans uses CNN models as a black-box, where the raw images are passed to the network for object detection. Considering the real-world scenarios where most of the baggage items are heavily occluded that even the human experts misses them, it will be very difficult for these frameworks to produce optimal results, because for a deep network to estimate the correct class, the feature kernels should be distinct. This condition is hard to fulfill for occluded objects obtained through raw images (without any initial processing) making thus the prediction of the true class quite a challenge. Note that this challenging aspect was also highlighted in [46], when they assessed the idea of using generative models to remove components that corresponds to the non-targeted objects in a recurrent fashion from the candidate scan. To alleviate this problem, the authors assumed that the generative models will only receive the supervision from the neighboring scales directly until the desired object is correctly classified. In [32], different CNN based object detectors were evaluated in on SIXray dataset (only SIXray10 subset was considered in [32] for detecting *guns* and *knives*). Despite the progress accomplished by the above works the challenge of correctly recognizing heavily occluded and cluttered items in SIXray dataset scans is still to be addressed.

## III. CONTRIBUTIONS

In this paper, we present a cascaded structure tensor (CST) framework through which each object within a scan (whether it is blurred, rotated or skewed) is automatically extracted and classified accordingly. The proposed framework is unique as it only uses a single feedforward CNN model for object recognition and instead of passing raw images or removing unwanted regions, the proposed framework intelligently extracts each object proposal by iteratively picking contour-based transitional information from different orientations within the candidate scan. The proposed framework is invariant to occlusion and can easily detect heavily cluttered objects as evident from the results section. The main contributions of the proposed framework are summarized below:

- This paper presents a novel object recognition framework which can extract objects from X-ray scans irrespective of their acquisition machinery and recognizes them using just a single feedforward CNN model.

- The proposed framework is invariant to the class imbalance problem since its trained directly on the balanced set of normal and suspicious items proposals rather than on the set of scans containing imbalanced ratio of normal and suspicious items.

- The extraction of object proposals in the proposed framework is performed through a novel CST framework, which analyzes the object transitions and coherency within a series of tensors generated from the candidate X-ray scan.

- The proposed CST framework exhibits high robustness to occlusion, scan type, noisy artifacts and to highly cluttered scenario.

- The proposed framework achieved mean intersection-over-union ($IoU$) score of 0.9644 and 0.9689, area under the curve ($AUC$) score of 0.9878 and 0.9950, and a mean average precision ($\mu_{AP}$) score of 0.9343 and 0.9595 for detecting normal and suspicious items from GDXray and SIXray dataset, respectively (see Section V).

- The proposed framework achieved 15.78% better runtime performance, compared to existing state-of-the-art solutions such as [29], [30], [31], [32], [42] and [44] which are based on exhaustive searches and anchor box estimations.

The rest of the paper is organized as follows: Section IV describes the proposed framework in detail, Section V presents the experimental setup, Section VI shows the results of the proposed framework and its comparison with state-of-the-art solutions, Section VII discusses the proposed framework in detail and Section VIII concludes the paper.

## IV. PROPOSED METHOD

Fig. 2 shows the block diagram of the proposed framework. The first step is a preprocessing step to enhance the contrast of the input image via an adaptive histogram equalization [47]. Afterwards, a series of tensors is generated where each tensor contains information about the targeted objects from different orientations. Using these tensors, the object proposals are automatically extracted and passed to the pre-trained ResNet architecture for object recognition. ResNet has extensively been used by the research community for object recognition due to its superior representation learning ability [48]. Each module within the proposed framework is described next:

### A. Preprocessing

The prime objective of the preprocessing stage is to enhance the low contrasted input. The contrast stretching in the proposed system is automatically performed through adaptive histogram equalization [47]. Let $\xi_X \in {}^{\sim MxN}$ the X-ray scan where $M$ and

$N$ denotes the number of rows and columns, respectively. Let $\wp$ be an arbitrary patch of $\xi_X$ where $\wp$ is obtained by dividing $\xi_X$ into $I \ x \ J$ grid of rectangular patches. The histogram of $\wp$ is computed and is locally normalized using the following relation:

$$\hbar = \Gamma\left(\frac{\Delta_\wp - \Delta_{\wp\min}}{M * N - \Delta_{\wp\min}}(L_M - 1)\right) \qquad (1)$$

where $\Delta_\wp$ is the cumulative distribution function of $\wp$, $\Delta_{\wp\min}$ is the minimum value of $\Delta_\wp$, $L_M$ represents the maximum grayscale level of $\xi_X$, $\Gamma(.)$ is the rounding function and $\hbar$ is enhanced histogram for $\wp$. This process is repeated for all the patches of $\xi_X$ to obtain the contrast stretched version $\xi_X{}^t$ as shown in Fig. 3. It can be observed that the occluded gun is clearly visible in the enhanced scan.

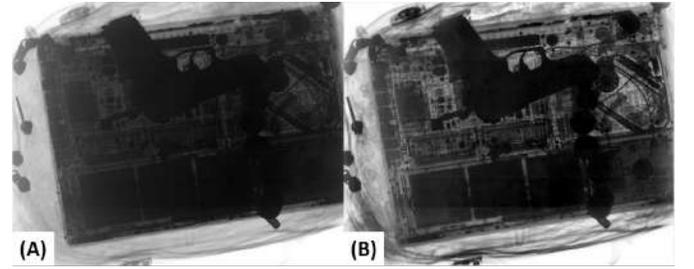

Fig. 3. Preprocessing stage: (A) original image containing occluded gun, (B) enhanced image $\xi_X{}^t$.

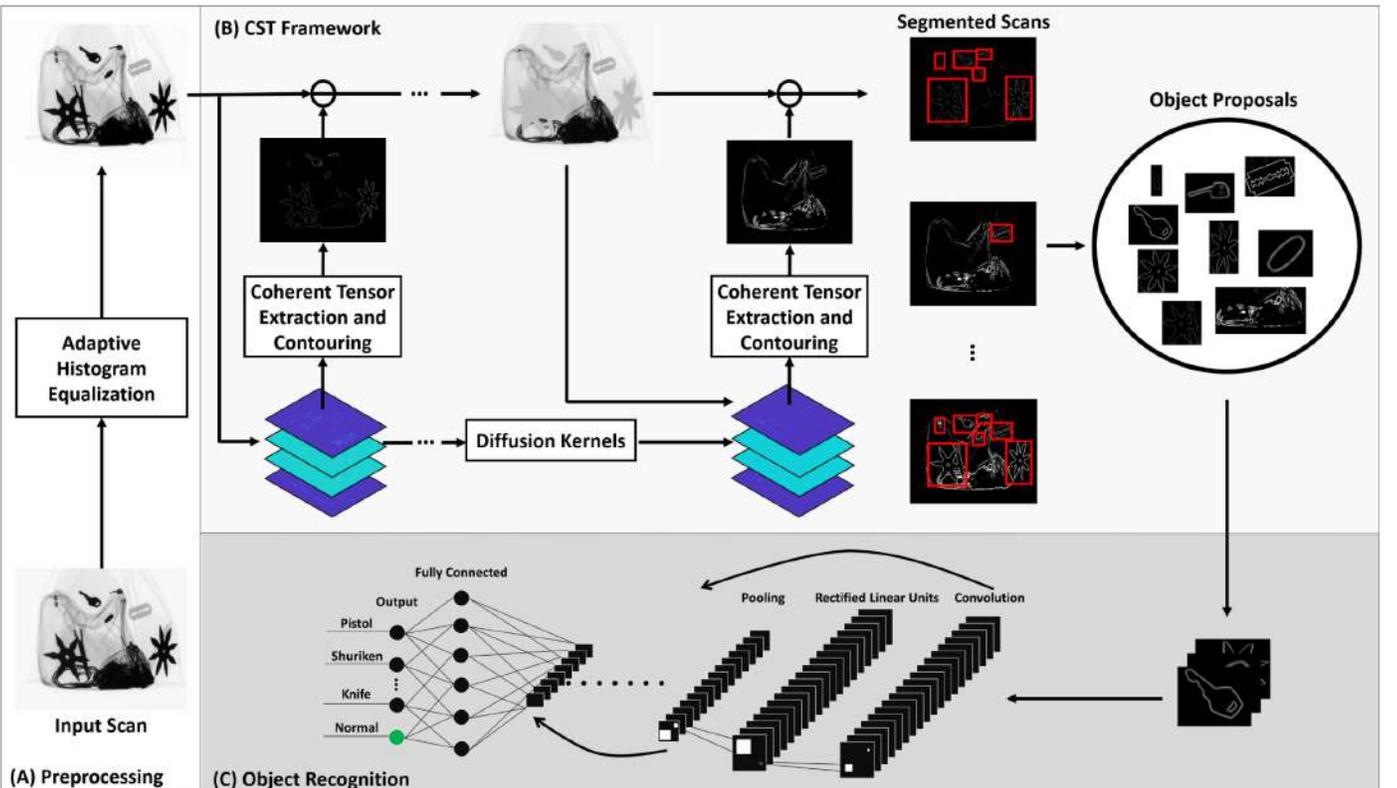

Fig. 2. Block diagram of proposed framework. The input scan is first preprocessed through (A). Afterwards, the proposal for each baggage item is automatically extracted through the CST framework (B). The extracted proposals are then passed to the pre-trained ResNet50 model for recognition (C).

## B. Cascaded Structure Tensor (CST) Framework

In order to extract the objects from the candidate scan, we propose a CST framework. CST is a segmentation framework based on structure tensors, a second moment matrix derived from the directional gradients of an image which reflects the predominant orientation of the local patterns within the input scan. In general, for $N$ number of orientations, the structure tensor $\delta_\Im$, in the proposed framework, generates $N^2$ coherent representations of the candidate image $\xi_X^t$ using the partial image gradients $\nabla \xi_X^t$ with respect to $\vartheta$ orientations where $\vartheta = \frac{2\pi k}{N}$ and $k$ is varied from 1 to $N$:

$$\delta_\Im = \begin{bmatrix} \Im_1^1(u,v) & \Im_2^1(u,v) & \cdots & \Im_g^1(u,v) \\ \Im_1^2(u,v) & \Im_2^2(u,v) & \cdots & \Im_g^2(u,v) \\ \vdots & \vdots & \ddots & \vdots \\ \Im_1^g(u,v) & \Im_2^g(u,v) & \cdots & \Im_g^g(u,v) \end{bmatrix} \quad (2)$$

where each tensor $\Im_j^i(u,v)$ is computed through:

$$\Im_j^i(u,v) = \sum_{u_i \in \omega_u} \sum_{v_j \in \omega_v} \varphi(u_i,v_j) \Im_i'(u-u_i,v-v_j) \Im_j'(u-u_i,v-v_j) \quad (3)$$

and

$$\Im_i'(u-u_i,v-v_j) = \frac{\partial \xi_X^t(u-u_i,v-v_j)}{\partial u} \quad (4)$$

$$\Im_j'(u-u_i,v-v_j) = \frac{\partial \xi_X^t(u-u_i,v-v_j)}{\partial v} \quad (5)$$

$\varphi(u,v)$ is a parametric Gaussian window that iteratively defuses the outliers within the gradients $\Im_i'$ and $\Im_j'$ during the computation of each tensor, $\omega_u$ and $\omega_v$ denotes the kernel width and height respectively. Afterwards, the tensor with the maximum coherency is automatically selected by evaluating the set of eigenvalues. Since $\Im_j^i(u,v)$ is a non-square matrix, the eigenvalues are obtained indirectly through singular value decomposition (SVD). Let $\Im \in \sim^{MxN}$ be the non-square tensor of the candidate input $\xi_X^t$. The SVD representation $\Im_{SVD}$ of $\Im$ can be obtained through:

$$\Im_{SVD} = \Upsilon V^* = U \sum V^* \quad (6)$$

$$\Upsilon = U \sum = \begin{bmatrix} U_{11} & U_{12} & \cdots & U_{1M} \\ U_{21} & U_{22} & \cdots & U_{2M} \\ \vdots & \vdots & \ddots & \vdots \\ U_{M1} & U_{M2} & \cdots & U_{M^2} \end{bmatrix} \begin{bmatrix} \Sigma_{11} & \Sigma_{12} & \Sigma_{13} & \cdots & \Sigma_{1N} \\ \Sigma_{21} & \Sigma_{22} & \Sigma_{23} & \cdots & \Sigma_{2N} \\ \vdots & \vdots & \vdots & \ddots & \vdots \\ \Sigma_{M1} & \Sigma_{M2} & \Sigma_{M3} & \cdots & \Sigma_{MN} \end{bmatrix} \quad (7)$$

$$\Im_{SVD} = \begin{bmatrix} \Upsilon_{11} & \Upsilon_{12} & \Upsilon_{13} & \cdots & \Upsilon_{1N} \\ \Upsilon_{21} & \Upsilon_{22} & \Upsilon_{23} & \cdots & \Upsilon_{2N} \\ \vdots & \vdots & \vdots & \ddots & \vdots \\ \Upsilon_{M1} & \Upsilon_{M2} & \Upsilon_{M3} & \cdots & \Upsilon_{MN} \end{bmatrix} \begin{bmatrix} V_{11}^* & V_{12}^* & \cdots & V_{1N}^* \\ V_{21}^* & V_{22}^* & \cdots & V_{2N}^* \\ \vdots & \vdots & \ddots & \vdots \\ V_{N1}^* & V_{N1}^* & \cdots & V_{N^2}^* \end{bmatrix} \quad (8)$$

where $U \notin \sim^{MxM}$ is a unitary matrix such that $UU^* = U^*U = I_M$ where $U^*$ is a conjugate transpose of $U$, $V^*$ is a conjugate transpose of a unitary matrix $V \notin \sim^{NxN}$ such that $VV^* = V^*V = I_N$, $\sum \notin \sim^{MxN}$ is a non-square diagonal matrix containing singular values of $\Im$, $I_M$ and $I_N$ are the $MxM$ and $NxN$ identity matrices. The eigenvalues of $\Im$ are then obtained through:

$$\Im \Im^* = \Im_{SVD} \Im_{SVD}^* = U \sum V^* V \sum^* U^* = U(\sum \sum^*) U^* \quad (9)$$

where $U$ contains the eigenvectors corresponding to $\Im \Im^*$ and the non-zero values of $\sum$ are the square root of eigenvalues of $\Im$. The tensor with the maximum coherency $\Im_\Theta$ is the one with the maximum eigenvalue strength. Therefore, by analyzing the extracted eigenvalues of each tensor, the one with the maximum coherency is automatically selected.

### 1) Extraction of Object Proposals

After obtaining the highly coherent representation of the candidate scan $\Im_\Theta$, it is binarized and morphologically enhanced to remove the unwanted blobs and noisy artifacts. Then, object contours are extracted from it, based upon the strength of their transitions with respect to the background. The purpose of computing contours over edges here is to obtain the isolated closed representation of the objects within the scan. Afterwards, each object is labeled through connected component analysis and for each labelled object, a bounding box is generated based upon minimum bounding rectangle technique that analyzes the minimum and maximum pixel values in both image dimensions. This bounding box is then used in the extraction of the respective object proposal from the candidate image. Moreover, the extracted object proposals are then removed from the candidate scan so that the CST framework can pick the transitions of the remaining objects in the next iteration. This process is repeated until there are no more objects to extract within the candidate scan. The detailed pseudocode of the proposed CST framework is presented below:

| **Algorithm:** Proposed CST Framework |
| --- |
| **Input:** Enhanced Image $\xi_X^t$, Number of Orientations $N$ |
| **Output:** Object Proposals $O_\Upsilon^\rho$ |
| 1:   $O_\Upsilon^\rho \leftarrow \phi$ |
| 2:   $\varsigma_f \leftarrow \phi$ |
| 3:   hasObjects $\leftarrow$ true |
| 4:   *while* hasObjects *is true*, *do* |
| 5:     $[\Im_1^i, \Im_2^i, \ldots, \Im_g^g] \leftarrow computeTensors(\xi_X^t, N, \varsigma_f)$ |

6:    $\mathfrak{I}_\Theta \leftarrow getCoherentOne([\ \mathfrak{I}_1^1, \mathfrak{I}_2^1, \cdots, \mathfrak{I}_g^g])$

7:    $\mathfrak{I}_\Theta^\Sigma \leftarrow removeBlobs(\ \mathfrak{I}_\Theta)$

8:    $\Phi_\Theta \leftarrow computeContours(\ \mathfrak{I}_\Theta^\Sigma)$

9:    $\ell_\Phi \leftarrow labelObjects(\ \Phi_\Theta)$

10:   **foreach** $\ell \in \ell_\Phi$, **do**

11:    $\Phi_\ell \leftarrow getLabelImage(\ \ell, \Phi_\Theta)$

12:    $[r, c] \leftarrow size(\Phi_\ell)$

13:    $\gamma_{min} \leftarrow \underset{i \in r}{\arg\min}(find(\Phi_{\ell,i} \neq 0, 1, 'first'))$

14:    $\gamma_{max} \leftarrow \underset{i \in r}{\arg\max}(find(\Phi_{\ell,i} \neq 0, 1, 'last'))$

15:    $\Phi_\ell \leftarrow \Phi_\ell{}'$ // transpose

16:    $\chi_{min} \leftarrow \underset{i \in c}{\arg\min}(find(\Phi_{\ell,i} \neq 0, 1, 'first'))$

17:    $\chi_{max} \leftarrow \underset{i \in c}{\arg\max}(find(\Phi_{\ell,i} \neq 0, 1, 'last'))$

18:    $w \leftarrow \chi_{max} - \chi_{min}$

19:    $h \leftarrow \gamma_{max} - \gamma_{min}$

20:    $\beta_B \leftarrow [\ \chi_{min}, \gamma_{min}, w, h]$

21:    $O^\rho \leftarrow cropImage(\ \Phi_\Theta, \beta_B)$

22:    $O_\Upsilon^\rho \leftarrow Append(\ O_\Upsilon^\rho, O^\rho)$

23:   **end**

24:   $\xi_X^t \leftarrow removeObjects(\ \xi_X^t, \Phi_\Theta)$

25:   $\varsigma_f \leftarrow updateScalingFactor(\ \varsigma_f)$

26:   hasObjects $\leftarrow checkMoreObjects(\ \xi_X^t)$

27: **end**

28: return $O_\Upsilon^\rho$

Fig. 4 shows a single pass of CST framework in which a tensor with the maximum coherency is generated for the candidate scan, highlighting the prominent objects along the orthogonal orientations with $N = 4$.

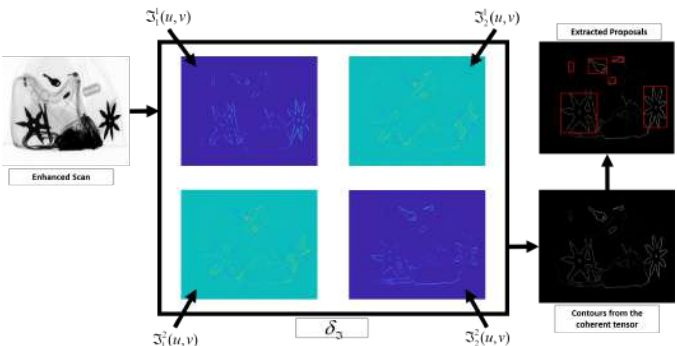

Fig. 4. Single pass of CST framework for orthogonal orientations

### C. Object Recognition

After extracting the object proposals, they are passed to a deep CNN model for recognition. For this purpose, we have used a pre-trained ResNet50 model as it is extensively used by the research community due to its good performance in catering the vanishing gradient problem through the residual blocks [48]. The blending of CST and ResNet50 model makes our framework exhibits attractive features. In terms of efficiency, the proposed framework is based on a single feedforward network with no iterative searches or anchor box estimations, so it requires less training time and provides best results for detecting heavily occluded and cluttered objects. As will be evidenced in the experiments it is faster compared to the state-of-the-art one staged and two staged architectures [8-10]. The detailed architectural description and hyper-parameters of ResNet50 is shown in Table I below:

TABLE I
Architectural description and hyperparameters of pre-trained ResNet50

| Layers | Description |
|---|---|
| Input Layer | Minimum resolution: 224x224x3 |
| Convolution Layers | 7x7x3 with stride [1, 1] and padding [3, 3, 3, 3] |
| | 1x1x64 with stride [1, 1] and padding [0, 0, 0, 0] |
| | 3x3x64 with stride [1, 1] and same padding |
| | 1x1x256 with stride [1, 1] and padding [0, 0, 0, 0] |
| | 3x3x128 with stride [1, 1] and same padding |
| | 1x1x128 with stride [1, 1] and padding [0, 0, 0, 0] |
| | 1x1x512 with stride [1, 1] and padding [0, 0, 0, 0] |
| | 3x3x256 with stride [1, 1] and same padding |
| | 1x1x1024 with stride [1, 1] and padding [0, 0, 0, 0] |
| | 1x1x2048 with stride [1, 1] and padding [0, 0, 0, 0] |
| | 3x3x512 with stride [1, 1] and same padding |
| Batch Normalization | Cross channel normalization |
| Rectified Linear Units (ReLU) | ReLU activation function |
| Pooling | Average Pooling: 7x7 with stride [7, 7] and padding [0, 0, 0, 0] |
| | Max Pooling: 3x3 with stride [2, 2] and padding [0, 0, 0, 0] |
| Addition Layers | Element-wise additions |
| Fully Connected Layer | 1000 fully connected layers |
| Softmax Layer | Based on softmax function |
| Classification Layer | Fine-tuned for the proposed framework |
| Optimizer | Stochastic gradient descent with momentum (SGDM) |

## V. EXPERIMENTAL SETUP

The proposed framework is evaluated against state-of-the art frameworks on different publicly available datasets using a variety of evaluation metrics. In this section, we have listed the detailed description of all the datasets and the evaluation metrics on which the proposed system is evaluated. Furthermore, this section also describes the training details of the pre-trained CNN model.

### A. Datasets

To evaluate the performance of the proposed model, we use GDXray [51] and SIXray [46] datasets, each of which is explained below:

#### 1) GRIMA X-ray Database

The GRIMA X-ray Database (GDXray) [49] is one of the oldest publicly available X-ray testing datasets made available for the computer vision community for research and educational purposes [48]. GDXray contains 19,407 X-ray scans arranged in welds, casting, baggage, nature and settings categories. In this paper, we only use the baggage scans to test the proposed framework as this is the only relevant category for suspicious items detection. The baggage group has 8,150 X-ray scans containing both occluded and non-occluded items. Apart from this, it contains the marked ground truths for *handguns*, *razor blades*, *shuriken* and *knives*. Furthermore, for more in-depth evaluation of the proposed framework, we have locally identified items as shown in Table II. Also, the original *handgun* category is broken down into *pistol* and *revolver* in order to see how efficiently the proposed system differentiates between the proposals for both items. The training and testing split on the GDXray dataset is made accordance with the standard defined in [49] i.e. 400 scans from B0049, B0050 and B0051 series containing proposals for *revolver*, *shuriken* and *razor blades*, respectively are used for training. This indicates that 5% scans were used for training while 95% scans were used for testing. To train our model for the identification of other items, we used the same percentage as indicated in Table II.

### 2) Security Inspection X-ray Dataset

Security Inspection X-ray (SIXray) [46] is one of the largest datasets for the detection of heavily occluded and cluttered suspicious items. It contains 1,059,231 color X-ray scans having 8,929 suspicious items which are classified into six groups i.e. *gun*, *knife*, *wrench*, *plier*, *scissor* and *hammer*. All the images are stored in JPEG format and the detailed description of the dataset is presented in Table III. To validate the performance of the proposed framework against the class imbalance problem, the same subsets have been utilized, as described in [46], in which the ratio of suspicious items and normal objects have been matched with real-world scenarios. Also, the ratio of 4 to 1 for training and testing has been maintained in accordance with [46]. Apart from this, the complete dataset has been annotated by experts [46], and these annotations served as a ground truth for the validation of the framework. Also, for both GDXray and SIXray datasets, we have added a separate *normal* class to filter the proposals of miscellaneous and unimportant items like keys and bag zippers etc., which are generated by CST framework. The *normal* class is not considered in the evaluations since it is only added to prevent the misclassification of such miscellaneous items as suspicious. Moreover, it should also be noted from Table II and III that for each dataset we have trained the classification model on the balanced set of normal and suspicious items proposals where the excessive normal items proposals are discarded to avoid the classifier biasness. Also, it is evident that the ratio of proposals in the GDXray and SIXray datasets is extremely low. So, in order to effectively train the generalized ResNet50 model, we further dropped the proposals from SIXray dataset to make a balanced training set as shown in Table IV.

TABLE IV
CONFIGURATION OF PROPOSALS FOR TRAINING RESNET50 ON BOTH DATASETS

| Datasets | Training Proposals | |
|---|---|---|
| | Normal | Suspicious |
| GDXray | 28049 | 28053 |
| SIXray | 28082 | 28093 |

### B. Training Details

The classification of baggage items within the proposed framework is performed through pre-trained ResNet50 model after fine-tuning it on the object proposals extracted from the scans of GDXray and SIXray datasets.

TABLE II
DETAILED DESCRIPTION OF GDXRAY (BAGGAGE) DATASET

| Acquisition Machine | Categories | Total Scans | Dataset Split | Training Proposals | Items |
|---|---|---|---|---|---|
| LASER scanner LS85 SDR, Image Intensifier, X-ray emitter (Poksom PXM-20BT) and detector (Cannon CXDI-50G) | Baggage | 8,150 | Training: 788 scans# <br> Testing: 7,362 scans | Total Proposals: 140,264 <br> Normal Considered: 28049 <br> Suspicious Considered: 28053 <br> Normal Discarded: 84,162 <br> **Average Proposals: 178 per scan** | • Pistol** <br> • Revolver** <br> • Shuriken <br> • Knife <br> • Razor blades <br> • Chip* <br> • Mobile* |

\* These items have been identified locally for the more in-depth validation of the proposed framework. Chip class represents all the electronic gadgets including laptops (except mobile phones).
\*\* Original *handgun* category is further broken down into *pistol* and *revolver* because both items are found to be in abundance within the dataset.
# 400 scans from B0049, B0050 and B0051 series are used for extracting *revolver*, *shuriken* and *razor blades* as per the criteria defined in [49] and using this same percentage of training and testing split, we used 388 more scans to train the model for the extraction of other items.

TABLE III
DETAILED DESCRIPTION OF SIXRAY DATASET

| Acquisition Machine | Categories | Total Scans | Dataset Split | Training Proposals | Items |
|---|---|---|---|---|---|
| Nuctech Dual-energy X-ray Scanner | SIXray10 | 98,219 | **Training: 78,575 scans** <br> **Testing: 19,644 scans** | Total Proposals: 12,179,125 <br> Normal Considered: 2,435,819 <br> Suspicious Considered: 2,435,825 <br> Normal Discarded: 7,307,481 | • Gun <br> • Knife <br> • Wrench <br> • Plier <br> • Scissor <br> • Hammer |
| | SIXray100 | 901,829 | **Training: 721,463 scans** <br> **Testing: 180,366 scans** | Total Proposals: 111,826,765 <br> Normal Considered: 22,365,348 <br> Suspicious Considered: 22,365,353 <br> Normal Discarded: 67,096,064 | |
| | SIXray1000 | 1,051,302 | **Training: 841,042 scans** <br> **Testing: 210,260 scans** | Total Proposals: 130,361,510 <br> Normal Considered: 26,072,296 <br> Suspicious Considered: 26,072,302 <br> Normal Discarded: 78,216,912 <br> **Average Proposals: 155 per scan** | |
| Total Scans | 1,059,231 | | Positives | 8,929 scans | |

| Negatives | 1,050,302 scans |

The training process was conducted for 30 epochs having a mini-batch size of 1 using MATLAB R2019a with deep learning toolbox, on a machine with an Intel Core i5-8400@2.8GHz processor, 16 GB RAM and NVIDIA RTX 2080 GPU with CUDA 7.5. The optimization during the training phase was performed through SGDM [50]. The base learning rate is chosen empirically as 0.001 with a piecewise training scheduler having the decaying factor of 0.5 every 2 epochs and a momentum of 0.9. Furthermore, the cross-entropy loss function $\zeta_L$ is employed during training that is computed through Eq. (10):

$$\zeta_L = -\sum_{i=1}^{s}\sum_{j=1}^{\omega} y_{i,j}\log(p_{i,j}) \qquad (10)$$

where $s$ denotes the total number of samples, $\omega$ denotes total number of classes, $y_{i,j}$ is a binary indicator stating whether $i^{th}$ sample belongs to $j^{th}$ class and $p_{i,j}$ is the predicted probability of the $i^{th}$ sample for $j^{th}$ class. Fig. 5 shows the training performance of the proposed framework based on the pre-trained ResNet50.

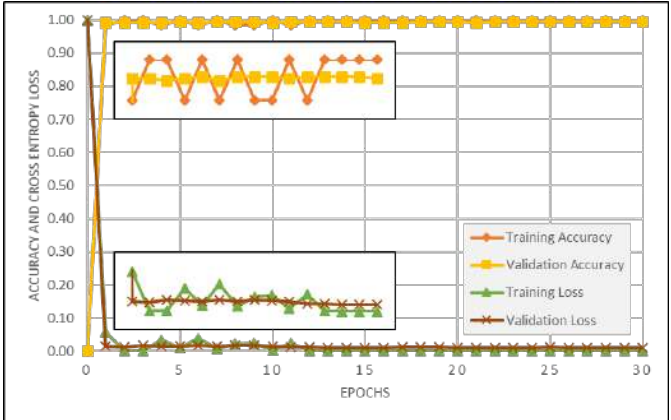

Fig. 5. Training performance of the proposed framework. The first 15 epochs are also zoomed for better visualization (loss and accuracy are normalized here for better visualization).

The training of the ResNet50 architecture in the proposed framework was conducted only once on the combination of objects proposals extracted from the scans of both GDXray and SIXray datasets.

### C. Evaluation Criteria

The performance is evaluated based on the following metrics:

#### 1) Intersection over Union

Intersection over Union ($IoU$) describes the overlapping area between the extracted object bounding box and the corresponding ground truth. It is also known as the Jaccard's similarity index and it is computed through Eq. (11):

$$IoU = \frac{\Lambda(\beta_B \bigcap G_\tau)}{\Lambda(\beta_B \bigcup G_\tau)} \qquad (11)$$

where $\beta_B$ is the extracted bounding box for the object, $G_\tau$ is the ground truth and $\Lambda(.)$ computes the area of the passed region. Although, $IoU$ measures the ability of the proposed framework to extract the corresponding object, it does not measure the detection capabilities of the proposed framework.

#### 2) Accuracy

Accuracy describes the performance of the proposed framework to correctly identify an object and no-object regions as described in Eq. (12) below:

$$Accuracy = \frac{T_P + T_N}{T_P + F_N + T_N + F_P} \qquad (12)$$

where $T_P$ denotes the true positive samples, $T_N$ denotes the true negative samples, $F_P$ denotes the false positive samples and $F_N$ denotes the false negative samples. Note that if $F_P = 0$ and $F_N = 0$ then $Accuracy = 100\%$.

#### 3) Recall

Recall, or sensitivity, is the true positive rate ($T_{PR}$) which indicates the completeness of the proposed framework to correctly classifying the object regions. It is computed through Eq. (13):

$$T_{PR} = \frac{T_P}{T_P + F_N} \qquad (13)$$

Note that if $F_N = 0$ then $T_{PR} = 100\%$.

#### 4) Precision

Precision ($P_R$) describes the purity of the proposed framework in correctly identifying the object regions against the ground truth and it is computed through Eq. (14).

$$P_R = \frac{T_P}{T_P + F_P} \qquad (14)$$

Note that if $F_P = 0$ then $P_R = 100\%$.

#### 5) Average Precision

Average precision represents the area under the precision-recall ($P_{RC}$) curve. It is a measure that indicates the ability of the proposed framework to correctly identifying positive samples (object proposals in our case) of each class/ group. The $P_{RC}$ curve for each class is generated by varying the classification threshold from 0 to 1 in steps of 0.001 and computing the recall and precision value for the respective class in each iteration through Eq. (13) and (14). Afterwards, the average precision is computed through:

$$\lambda_P = \int_a^b P_R(r)dr \approx \left(\sum_{\upsilon=a}^{b} P_R(\upsilon)\nabla T_{PR}(\upsilon)\right) \qquad (15)$$

where $\lambda_P$ denotes the average precision, $\nabla T_{PR}$ represents the change (difference) in the consecutive recall values, $a$ is the starting point of the integration interval which is equal to 0, and $b$ is end point of the integration interval which is 1. After computing the $\lambda_P$ for each class, the $\mu_{AP}$ score is computed using Eq. (16):

$$\mu_{AP} = \frac{1}{\varepsilon}\left(\sum_{k=0}^{\varepsilon-1}\lambda_P(k)\right) \tag{16}$$

where $\varepsilon$ denotes the number of classes in each respective dataset.

*6) $F_1$ Score:*

$F_1$ score measures the ability of any model that how well it can correctly classified samples in a highly imbalanced scenario. It is computed by considering both $P_R$ and $T_{PR}$ scores using Eq. (17):

$$F_1 = \frac{2 \times (Precision \times Recall)}{(Precision + Recall)} \tag{17}$$

*7) Receiver Operator Characteristics (ROC) Curve:*

In order to further validate the proposed framework for object recognition, ROC curves are computed. ROC curves indicate the degree of how much the proposed framework confuses between different classes. It is computed by varying the classification threshold from 0 to 1 in steps of 0.001 and computing the $T_{PR}$ and the false positive rate ($F_{PR}$) using Eq. (13) and (18):

$$F_{PR} = 1 - Specificity = 1 - \frac{T_N}{T_N + F_P} = \frac{F_P}{T_N + F_P} \tag{18}$$

where specificity is the true negative rate. Note that if $T_N = 0$ then $F_{PR} = 100\%$. After computing the ROC curves for each class within GDXray and SIXray datasets, $AUC$ is computed by numerically integrating the ROC curve.

*8) Qualitative Evaluations*

Apart from the quantitative evaluations, the proposed framework is thoroughly validated through qualitative evaluations as shown in the results and the discussion sections.

## VI. Results

In this section we report the results obtained through a comprehensive series of experiments conducted with GDXray and SIXray datasets. In Subsection-A we present the performance of the proposed system evaluated based on the metrics explained in Section IV (C). Then in Subsection-B we report a comparative study with state-of-the-art methods.

### A. System Validation

Table V shows the mean $IoU$ ratings of the proposed framework on GDXray and SIXray datasets, where we can see that the proposed framework achieved the mean $IoU$ score of 0.9644 on GDXray and 0.9689 on SIXray dataset, respectively. In Fig. 6 and 7, we report some qualitative results for the GDXray and SIXray datasets, respectively. The depicted examples illustrate that our framework can effectively extract and localize suspicious items from the grayscale and color X-ray scans w.r.t their ground truths. Note that for the fair and direct comparison, only the originally identified items in each dataset are shown in

Fig. 6 and 7 (and not the locally identified items) along with their original ground truths. Fig. 8 and Fig. 9 reports other qualitative results of items exhibiting high occlusion and clutter from GDXray and SIXray dataset, respectively.

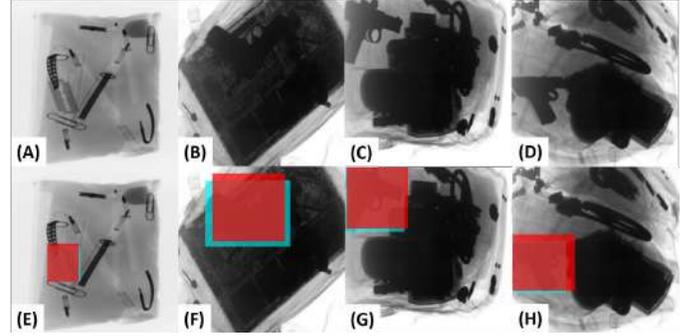

Fig. 6. GDXray dataset: Example of detected objects. Red color shows the extracted regions, while cyan color shows the ground truth. The figure illustrates the capacity of the proposed framework that how accurately it can extract the suspicious items w.r.t their ground truth.

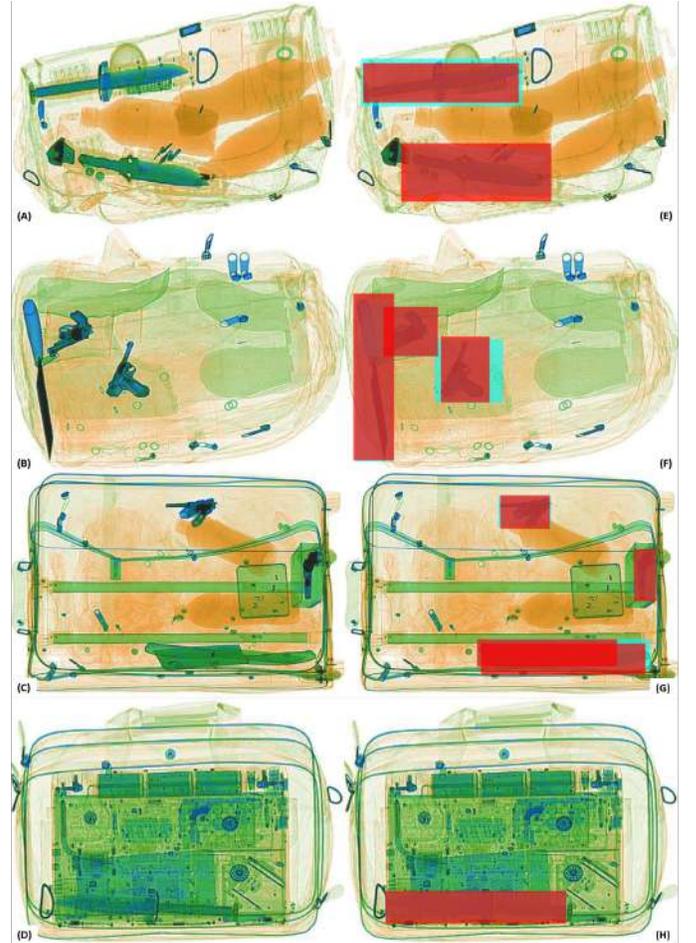

Fig. 7. SIXray dataset: Examples of detected objects. Red color depicts the extracted regions, while cyan color shows the ground truth. The left column shows the original scans. The figure illustrates the capacity of the proposed framework that how accurately it can extract the suspicious items w.r.t their ground truth.

Through these examples, we can appreciate the capacity of the framework for accurately extracting and recognizing items in a such challenging conditions. For example, in Fig. 8 (A) and (B), we can observe that the *chip* object has been severely occluded by the bundle of wires. However, the proposed framework, due to its adaptive contrast adjustment strategy and ability to analyze transitional patterns, has effectively recognized the occluded *chip* object as evident from Fig. 8 (D) and (E). Similar example can also be seen in Fig. 8 (J), (K) and (L) where the proposed framework easily extracted the partially occluded *pistol* and *mobile* from Fig. 8 (G), (H) and (I).

| Items | GDXray | SIXray |
|---|---|---|
| Handguns (including pistol and revolver) | 0.9487 | 0.9811 |
| Knife) | 0.9872 | 0.9981 |
| Shuriken | 0.9658 | - |
| Chip | 0.9743 | - |
| Mobile | 0.9425 | - |
| Razor Blades | 0.9681 | - |
| Wrench | - | 0.9894 |
| Plier | - | 0.9637 |
| Scissor | - | 0.9458 |
| Hammer | - | 0.9354 |
| **Mean ± STD** | **0.9644 ± 0.0165** | **0.9689 ± 0.0249** |

For the SIXray dataset, we can also see that how well the proposed framework has recognized the partially and heavily occluded suspicious items especially from Fig. 9 (B), (D), (F), (J), (N) and (P). It is also worth noting here that although in Fig. 9 (G) and (K), the intensity of *gun* or *knife* is similar. So, any model relying only on transitions would have considered them as one object (since they are merged together) However, the proposed framework has effectively discriminated between them due to its ability to measure the coherency between the transitions of similar intensity objects.

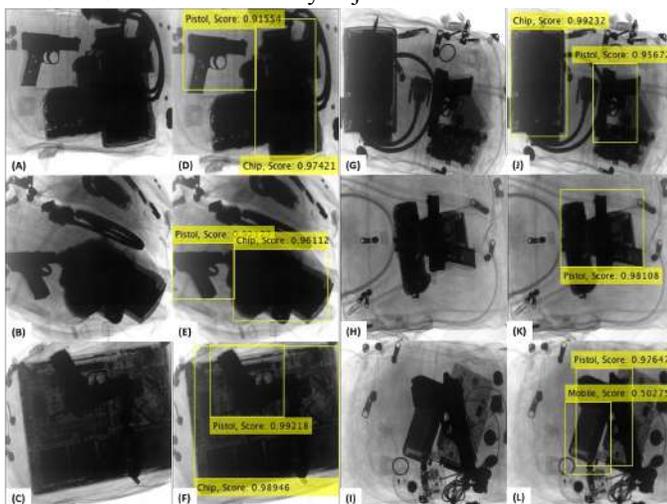

Fig. 8. Performance of proposed framework in recognizing heavily occluded and cluttered items from GDXray. 1st and 3rd column shows the original scans. It is extremely difficult for a conventional deep object detector (or even a human expert) to recognize baggage items from low-textured grayscale X-ray scans such as these but the proposed framework, due to its ability to analyze transitional patterns and generate contour-based proposals, can efficiently recognize these items.

Fig. 10 reports the validation of our system through the ROC curves. It should be noted here that the true positives represent pixels of the items which are correctly identified and true negatives represent the pixels of the background which are correctly identified. Furthermore, the reported scores are computed based upon the performance of the proposed framework for both the correct extraction and recognition of the suspicious items. For example, if the item has been correctly extracted by the CST framework but it has been misclassified by the ResNet50, then we counted that as a false negative in the scoring. It can be observed from Fig. 10 that the minimum *AUC* score is achieved for *Razor Blades* (i.e. 0.9582). Here, another reason of achieving lowest score for the *Razor Blades* is that the intensity differences between the *Razor Blades* and the background is very minimum within the scans of GDXray dataset, due to which sometimes it gets missed by the CST framework.

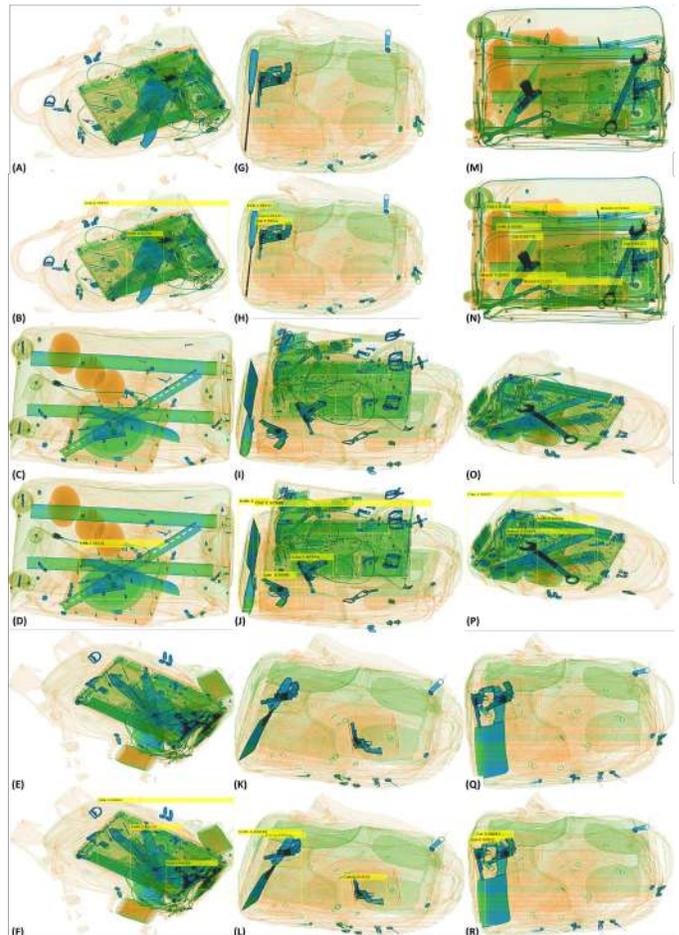

Fig. 9. Performance of the proposed framework in recognizing objects from SIXray. 1st, 2nd and 3rd row shows the original scans. It can be observed that the proposed framework can efficiently recognize heavily occluded, cluttered and concealed items from the challenging SIXray dataset scans.

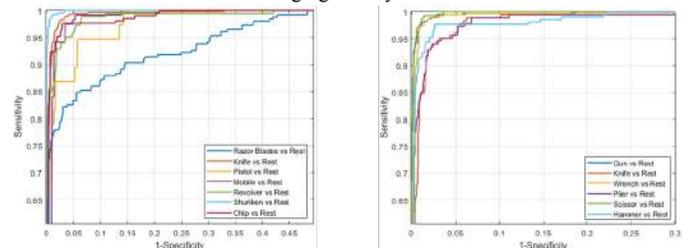

Fig. 10. ROC curves for items recognition on GDXray (left) and SIXray (right). The proposed framework achieves good performance for classifying normal and suspicious items from both datasets where it achieves the minimum *AUC* score of 0.9582 for identifying *Razor Blades*.

Fig. 11 shows the $P_{RC}$ curves computed on the GDXray and SIXray datasets. It can be observed from Fig. 11 that the proposed framework is extremely robust in detecting occluded and heavily cluttered suspicious items. Furthermore, Table VI shows a performance comparison of the proposed framework in terms of $\lambda_P$ for each class and $\mu_{AP}$ for recognizing heavily occluded baggage items from the GDXray and SIXray scans. It can be observed from Table VI that the proposed framework achieved the $\mu_{AP}$ score of 0.9343 and 0.9595 GDXray and SIXray dataset, respectively. Moreover, the $\mu_{AP}$ score achieved by the proposed framework for the *handgun* category on GDXray dataset is 0.9101 (which is the average of *pistol* and *revolver* class). Apart from this, the proposed framework is further validated by its generalization ability to recognize objects from both datasets through the $AUC$ scores, as shown in Table VII, where it can be observed that the proposed framework is extremely robust in recognizing objects from the challenging SIXray dataset [46] and achieved the $AUC$ score of 0.9878 and 0.9950 on GDXray and SIXray dataset, respectively.

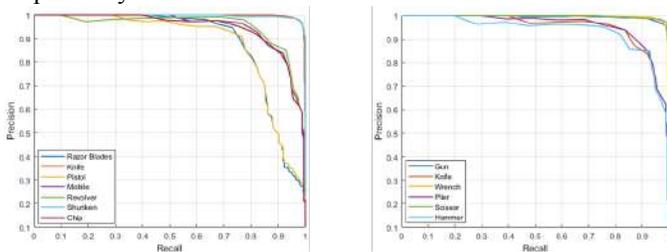

Fig. 11. Precision-recall curve for object recognition on GDXray (left) and SIXray (right). The proposed framework achieves the minimum $\mu_{AP}$ score of 0.8826 on GDXray and 0.9189 on SIXray for classifying *Razor Blades* and *Hammer*, respectively.

TABLE VI

$\mu_{AP}$ SCORES ON GDXRAY AND SIXRAY FOR THE DETECTION OF DIFFERENT SUSPICIOUS ITEMS. '-' INDICATES THAT THE RESPECTIVE ITEM IS NOT PRESENT IN THE DATASET

| Items | GDXray | SIXray |
|---|---|---|
| Razor Blades | 0.8826 | - |
| Knife | 0.9945 | 0.9347 |
| Pistol | 0.8762 | - |
| Mobile | 0.9357 | - |
| Revolver | 0.9441 | - |
| Shuriken | 0.9917 | - |
| Chip | 0.9398 | - |
| Gun | 0.9101* | 0.9911 |
| Wrench | - | 0.9915 |
| Plier | - | 0.9267 |
| Scissor | - | 0.9938 |
| Hammer | - | 0.9189 |
| **Mean ± STD** | **0.9343 ± 0.0442** | **0.9595 ± 0.0362** |

\* the $\lambda_P$ score of 'Gun' for GDXray dataset is the average of *pistol* and *revolver*.

TABLE VII

$AUC$ SCORES ON GDXRAY AND SIXRAY

| Items | GDXray | SIXray |
|---|---|---|
| Razor Blades | 0.9582 | - |
| Knife | 0.9972 | 0.9981 |
| Pistol | 0.9834 | - |
| Mobile | 0.9914 | - |
| Revolver | 0.9932 | - |
| Shuriken | 0.9987 | - |
| Chip | 0.9925 | - |
| Gun | 0.9883* | 0.9910 |
| Wrench | - | 0.9971 |
| Plier | - | 0.9917 |
| Scissor | - | 0.9990 |
| Hammer | - | 0.9932 |
| **Mean ± STD** | **0.9878 ± 0.0120** | **0.9950 ± 0.0031** |

\* the $AUC$ score of 'Gun' for GDXray dataset is the average of $AUC$ scores of pistol and revolver

### B. Comparative Study

For the GDXray dataset, we compared our framework with the methods [39], [42], [52] and [53] as shown in Table VIII. Contrary to these methods, we accessed our framework for all the performance criteria described in Section V (C). The performance comparison is nevertheless indirect as the experiment protocol in each study differs, where we (as well as authors in [39]) followed the standards laid in [49] i.e. we considered 400 images for training i.e. 100 for *razor blades*, 100 for *shuriken* and 200 for *handguns*. However, [39] used 600 images for testing purposes (200 for each item) and considered only 3 items whereas we considered 7 items and used 7,362 scans for testing. Apart from this, the authors in [42] considered a total of 3,669 selective images in their work having 1,329 *razor blades*, 822 *guns*, 540 *knives*, and 978 *shuriken*. To train Faster RCNN, YOLOv2 and Tiny YOLO models, they picked 1,223 images from the dataset and augmented them to generate 2,446 more images. The work reported in [53] involved 18 images only while [52] reports a study that is based on non-ML methods where the authors conducted 130 experiments to detect *razor blades* within the X-ray scans. It should be noted here that the proposed framework has been evaluated in the most restrictive conditions as compared to its competitors where the true positive samples (of the extracted items) were only counted towards the scoring when they were correctly classified by the ResNet50 model as well. So, if the item has been correctly extracted by the CST framework, if it was not correctly recognized by the ResNet50 model, we counted it as a misclassification for evaluation. Despite such restrictions, we were able to achieve 4.25% improvements in the precision, 2.97% improvements in the $F_1$ score as evident from Table VIII. Furthermore, our proposed framework achieved the $F_1$ score of 0.8025 when it was trained on scans from both GDXray and SIXray datasets. Although, the proposed framework lags from [42] in terms of accuracy but it has less significance because of the class imbalance problem here as the true negatives denote the actual background pixels which are in a very large quantity as compared to the pixels of the suspicious items (true positives). So, due to the accuracy paradox, the true metric of evaluation for the imbalanced classes is the $F_1$ score. Also, by observing the sensitivity of our proposed framework at the false positive rate of 0.35 in Table VIII, we can see that the proposed framework outperforms [39] by 1.2%. Similarly, for the SIXray dataset we compared our system with the methods proposed in [46] and [32] (the only two frameworks which have been applied on SIXray dataset till date). Note that these two works employed different pre-trained models, which we also reported in Table VIII for completeness. Moreover, for a direct and fair comparison with [46] and [32], we have trained the proposed framework on each subset of the SIXray dataset individually and reported the performance where *hammer* class is not considered in these experimentations as it was not considered in [46]. Note also that the SIXray dataset is divided into three subsets to address the problem of class imbalance. These subsets are named as SIXray10, SIXray100 and SIXray1000.

SIXray10 contains all 8,929 positive scans (having suspicious items) and 10 times the negative scans (which do not contain any suspicious item). Similarly, SIXray100 has all the positive scans and 100 times the negative scans. SIXray1000 contains only 1000 positive scans and all the negative scans (1,050,302 in total). So, the most challenging subset for class imbalance problem is SIXray1000. It is evident from Table IX that the proposed framework outperforms its competitors in terms of object classification and localization. Moreover, the performance comparison of proposed framework with [46] and [32] in recognizing individual items from each SIXray subset is shown in Fig. 12 where ResNet50 is used as a backbone. It can be observed from Fig. 12 that the proposed framework is very robust in detecting the suspicious baggage items from SIXray dataset scans. Also, it is worth noting here that the proposed framework achieved significant amount of improvements as compared to [46] for recognizing *wrench*, *plier* and *scissor* on SIXray10, SIXray100 and SIXray1000 subsets which further indicates that it's resistance to the class imbalance problem.

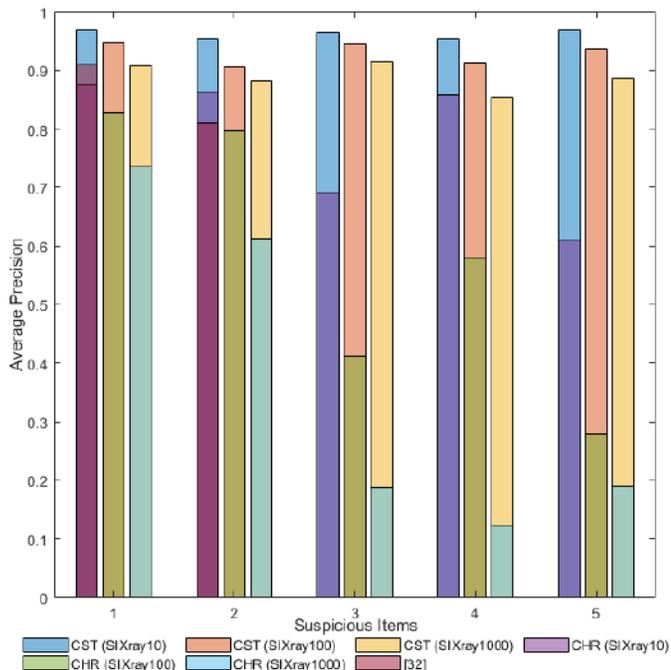

Fig. 12: Performance comparison of proposed framework with [46] and [32] in identifying suspicious items from SIXray subsets. ResNet50 is used as a backbone where "1" represents *gun*, "2" represents *knife*, "3" represents *wrench*, "4" represents *plier* and "5" represents *scissor* class.

In the last experiment we compared the computational performance of our system with standard one staged (such as YOLOv2, RetinaNet) and two staged detectors (such as Faster RCNN), as they have been widely used for suspicious items detection in the past. The results depicted in Table X shows that our system score the best average time performance in both training and testing outperforming in particular YOLOv2. Note that although YOLOv2 has significant improvements in computational performance over other two staged architectures, however it is developed for extracting objects from rich textured photographs. Also, due to its spatial constraints it does not detect small objects well.

## VII. Discussion

This paper presents a CST based deep learning framework that can identify heavily cluttered and occluded suspicious items from X-ray images. The proposed framework has been generalized to works on scans, irrespective of their type or acquisition properties. The proposed framework generates a series of object proposals using a novel CST segmentation scheme. These proposals are then passed to a pre-trained classification network for recognition. The proposed framework achieved a better time performance as compared to its competitors and it produces good results for the practical recognition of objects, irrespective of the scan texture or the object size and orientation. Since CST framework always look for the transitional information within the candidate scan so it's highly unlikely that it misses the proposal of any suspicious item as evident from Fig. 8 and 9. However, CST do generate proposals for the miscellaneous items like bag zippers etc. in large quality. But these proposals are filtered out by the ResNet50 model since we have trained it on the balanced set of normal and suspicious items proposal. Also, due to this factor, the proposed framework is invariant to the class imbalance problem and no matter how many negative or positive scans are fed to the proposed framework, it correctly classifies the normal and suspicious items even from the complex X-ray scans as evident from the evaluation metrics. In addition to this, the proposed framework efficiently extracts heavily cluttered and occluded objects, irrespective of the scan type as evident from Fig. 9. From the same figure, it can also be observed that the proposed framework correctly identified the *chip* (Fig. 9 (F), (J), (N) and (P)) from the SIXray dataset scan (although the bounding boxes are not exact due to segmentation error). We have rigorously tested the proposed framework on two publicly available datasets with different scan types, containing different objects and it outperformed the existing state-of-the-art solutions with various metrics, as evident from the results section. Also, the proposed CST framework is highly sensitive in picking merged, cluttered and overlapping items through its ability to measure the coherency of the transitions and by generating contour based proposals. In some rare scenarios, the proposed CST framework is limited, and unable to isolate close objects with similar intensity levels. For this reason, some noisy artifacts are sometimes observed in the object proposals which leads incorrect generation of bounding boxes (as it can be seen for chip class in Fig. 9). However, ResNet50 caters this segmentation error by correctly recognizing such object proposals because the contribution of these artifacts is very minimal in the overall proposal. Furthermore, the object recognition performance of the proposed framework is validated through the $\mu_{AP}$, $AUC$ and $F_1$ scores as shown in the results section. To further highlight the capacity of the proposed framework to detect occluded, non-occluded, clearly visible and highly concealed items from grayscale and color X-ray scans, we present some more qualitative results in Fig. 13 from which we can appreciate the robustness of the proposed system and its potential for deployment in the practical world for screening real-time threats on-the-fly. As stated above, the backbone of the proposed framework is the strength of transitions within the scans but from Fig. 13, we can clearly observe how effectively the low intensity *razor blades* have been extracted from the GDXray scans. Also, the ability of the proposed framework to discriminate between *revolver* and *pistol* can be easily seen in Fig. 13, whereas all the existing state-of-the-art solutions have considered it as a single *handgun* in their experimentations. For SIXray dataset, we can also see how effectively the proposed framework has extracted suspicious items from even the most complex x-ray scans (Fig. 13 last two columns).

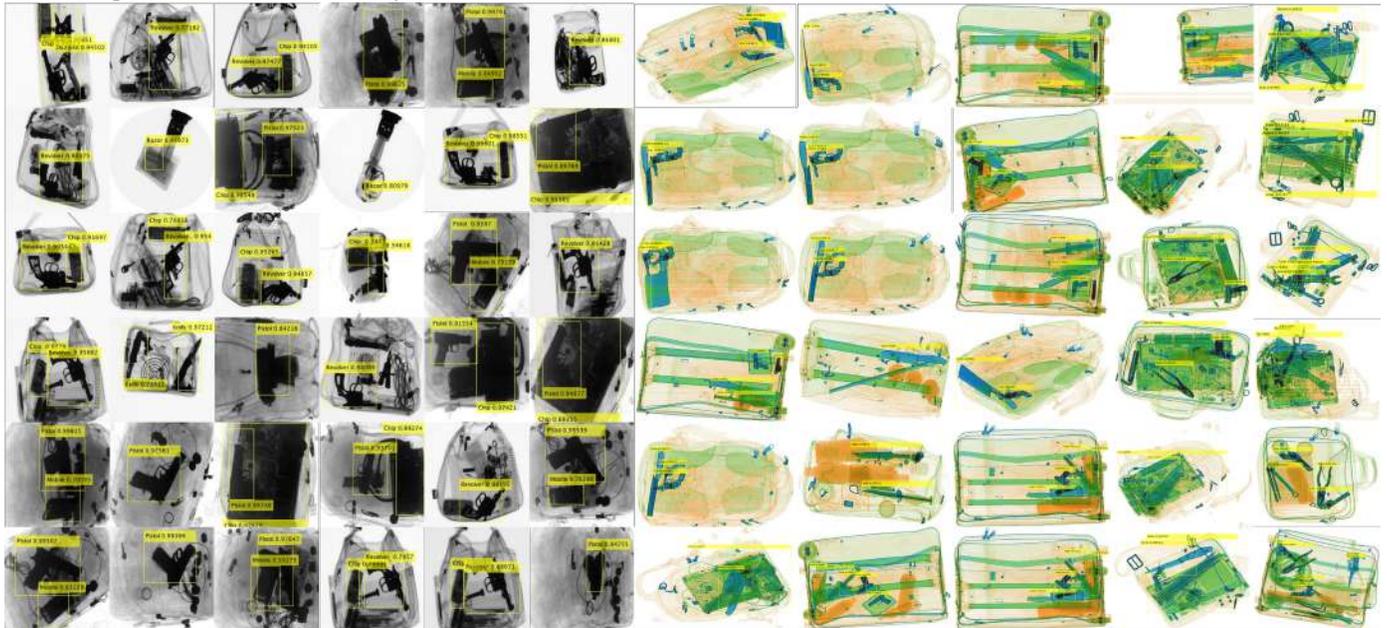

Fig. 13: Some examples from GDXray (left grid) and SIXray (right grid) dataset depicting the robust performance of proposed framework for detecting both clearly visible and extremely concealed items.

TABLE VIII
PERFORMANCE COMPARISON OF THE PROPOSED FRAMEWORK ON GDXRAY DATASET. BOLD INDICATES THE BEST SCORES WHILE THE SECOND-BEST SCORES ARE UNDERLINED. '-' INDICATES THAT THE METRIC HAS NOT BEEN COMPUTED. PROTOCOLS ARE DEFINED BELOW

| Criteria | Proposed (Generalized) | Proposed (GDXray) | Faster RCNN [42] | YOLOv2 [42] | Tiny YOLO [42] | AISM₁ [39]* | AISM₂ [39]* | SURF [39] | SIFT [39] | ISM [39] | [52] | [53] |
|---|---|---|---|---|---|---|---|---|---|---|---|---|
| Mean $AUC$ | 0.9878 | **0.9934** | - | - | - | <u>0.9917</u> | <u>0.9917</u> | 0.6162 | 0.9211 | 0.9553 | - | - |
| Accuracy | 0.9457 | <u>0.9833</u> | **0.9840** | 0.9710 | 0.89 | - | - | - | - | - | - | - |
| Sensitivity | 0.9428 | <u>0.9969</u> | 0.98 | 0.88 | 0.82 | **0.9975** | 0.9849 | 0.6564 | 0.8840 | 0.9237 | 0.89 | 0.943 |
| Specificity | <u>0.95</u> | **0.9650** | - | - | - | <u>0.95</u> | **0.9650** | 0.63 | 0.83 | 0.885 | - | 0.944 |
| False Positive Rate | <u>0.05</u> | **0.035** | - | - | - | <u>0.05</u> | **0.035** | 0.37 | 0.17 | 0.115 | - | 0.056 |
| Precision | 0.6985 | **0.9714** | <u>0.93</u> | 0.92 | 0.69 | - | - | - | - | - | 0.92 | - |
| $F_1$ Score | 0.8025 | **0.9836** | <u>0.9543</u> | 0.8996 | 0.7494 | - | - | - | - | - | 0.9048 | - |

Proposed (Generalized): Classes: 13, Split: *GDXray:* 5% for training and 95% for testing, *SIXray:* 80% for training and 20% for testing.
Proposed (GDXray): Classes: 7, Split: 5% for training and 95% for testing, Training Images: 400 (and 388 more for extra items), Testing Images: 7,362.
[39]: Classes: 3, Split: 40% for training, 60% for testing, Training Images: 400, Testing Images: 600 (200 for each category).
[42]: Classes: 4, Split: 80% for training and 20% for validation, Training Images: 3,669, Testing Images: 4.
[52]: Classes: 1, (non-ML approach).
[53]: Classes: 3, Total Images: 18.
* the ratings of AISM₁ and AISM₂ are obtained from the ROC curve of AISM for different $F_{PR}$ values.

TABLE IX
PERFORMANCE COMPARISON OF THE PROPOSED FRAMEWORK WITH EXISTING SOLUTIONS ON SIXRAY SUBSETS. BOLD INDICATES THE BEST SCORES WHILE THE SECOND-BEST SCORES ARE UNDERLINED.

| Criteria | Subset | ResNet50 + CST | ResNet50 [54] | ResNet50 + CHR [46] | DenseNet [55] | DenseNet + CHR [46] | Inceptionv3 [56] | Inceptionv3 + CHR [46] | [32] |
|---|---|---|---|---|---|---|---|---|---|
| Mean Average Precision | SIXray10 | **0.9612** | 0.7685 | 0.7794 | 0.7736 | 0.7956 | 0.7956 | 0.7949 | <u>0.86</u> |
| | SIXray100 | **0.9297** | 0.5222 | 0.5787 | 0.5715 | <u>0.5992</u> | 0.5609 | 0.5815 | - |
| | SIXray1000 | **0.8894** | 0.3390 | 0.3700 | 0.3928 | <u>0.4836</u> | 0.3867 | 0.4689 | - |
| | SIXray10 | **0.8254** | 0.5140 | 0.5485 | 0.6246 | <u>0.6562</u> | 0.6292 | 0.6354 | - |

| Localization Accuracy | SIXray100 | **0.7786** | 0.3405 | 0.4267 | 0.4470 | <u>0.5031</u> | 0.4591 | 0.4953 | - |
| | SIXray1000 | **0.7429** | 0.2669 | 0.3102 | 0.3461 | <u>0.4387</u> | 0.3026 | 0.3149 | - |

TABLE X

TIME PERFORMANCE COMPARISON OF THE PROPOSED FRAMEWORK WITH POPULAR OBJECT DETECTORS. ALL FRAMEWORKS HAVE BEEN EVALUATED USING PRE-TRAINED RESNET50 CLASSIFICATION NETWORK. BOLD INDICATES THE BEST SCORES WHILE THE SECOND-BEST SCORES ARE UNDERLINED.

| Machine Specifications | Average time performance in seconds | Proposed (ResNet50) | YOLOv2 (ResNet50) | RetinaNet-50 [11] | Faster R-CNN (ResNet50) | R-CNN (ResNet50) |
|---|---|---|---|---|---|---|
| Intel i5-8400@2.8GHz processor with 16 GB DDR3 RAM and NVIDIA RTX 2080 GPU | Training | **677.09 seconds** | <u>712.72 seconds</u> | 927.52 seconds | 19,600 seconds | 306,000 seconds |
| | Testing | **0.019 seconds per image** | <u>0.025 seconds per image</u> | 0.073 seconds per image | 0.55 seconds per image | 134.75 seconds per image |

## VIII. CONCLUSION

This paper presents a novel framework for the automated detection of normal and suspicious items from X-ray baggage scans. The proposed framework can extract and recognize heavily cluttered and occluded objects from X-ray scans irrespective of their type or acquisition machinery. The proposed system is rigorously tested on different publicly available datasets and is thoroughly compared with existing state-of-the-art solutions using different metrics. The proposed framework achieved the mean $IoU$ score of 0.9644 and 0.9689, $AUC$ score of 0.9878 and 0.9950, and a $\mu_{AP}$ score of 0.9343 and 0.9595 on GDXray and SIXray dataset, respectively for detecting heavily occluded and concealed items. The extraction of object proposals in the proposed framework is based on novel CST segmentation scheme, which generates a series of tensors and measure their coherency for accurately representing the transitional patterns. Furthermore, the proposed framework can detect multiple objects from X-ray scans, irrespective of the scan type and imbalanced classes. It is based only on a single feedforward CNN architecture and do not require any exhaustive searches and regression networks due to which it is more time efficient as compared to the popular two-staged CNN object detectors. The proposed system can be applied on normal photographs and popular large scale publicly available datasets for the autonomous object detection. This will be the object of future research.

## ACKNOWLEDGMENT

This work is funded by Center for Cyber-Physical System (C2PS), Khalifa University of Science and Technology, Abu Dhabi, United Arab Emirates.